\documentclass[sigconf,natbib=false]{acmart}

\usepackage[super]{nth}
\usepackage{caption}
\usepackage{subcaption}
\usepackage{scalerel,stackengine}
\stackMath
\newcommand\reallywidehat[1]{%
\savestack{\tmpbox}{\stretchto{%
  \scaleto{%
    \scalerel*[\widthof{\ensuremath{#1}}]{\kern.1pt\mathchar"0362\kern.1pt}%
    {\rule{0ex}{\textheight}}
  }{\textheight}%
}{2.4ex}}%
\stackon[-6.9pt]{#1}{\tmpbox}%
}

\newcommand{\remove}{\phantom}

\AtBeginDocument{%
  \providecommand\BibTeX{{%
    Bib\TeX}}}

\copyrightyear{2022}
\acmYear{2022}
\setcopyright{rightsretained}
\acmConference[SIGSPATIAL '22]{The 30th International Conference on Advances in Geographic Information Systems}{November 1--4, 2022}{Seattle, WA, USA}
\acmBooktitle{The 30th International Conference on Advances in Geographic Information Systems (SIGSPATIAL '22), November 1--4, 2022, Seattle, WA, USA}
\acmDOI{10.1145/3557915.3560968}
\acmISBN{978-1-4503-9529-8/22/11}

\RequirePackage[
  datamodel=acmdatamodel,
  style=acmnumeric,
  ]{biblatex}

\begin{document}

\title{A Novel Framework for Handling Sparse Data in Traffic Forecast}


\author{Nikolaos Zygouras}
\thanks{Part of this work was done while N. Zygouras was at the National and Kapodistrian University of Athens, Greece}

\affiliation{%
  \institution{Huawei Amsterdam Research Center}
  \country{Netherlands}
  }
\email{nikolas.zygouras@huawei.com}

\author{Dimitrios Gunopulos}
\affiliation{%
  \institution{National and Kapodistrian University of Athens}
  \country{Greece}}
\email{dg@di.uoa.gr}

\renewcommand{\shortauthors}{Zygouras et al.}

\begin{abstract}
      The ever increasing amount of GPS-equipped vehicles provides in real-time
    valuable traffic information for the roads traversed by the moving vehicles.
    In this way, a set of sparse and time evolving traffic reports is generated for each road.
    These time series are a valuable asset in order to forecast the future traffic condition.
    In this paper we present  a deep learning framework that encodes the sparse recent traffic information and forecasts the future traffic condition. 
    Our framework consists of a recurrent part and a decoder. The recurrent part employs an attention mechanism that encodes the traffic reports that are available at a particular time window. The decoder is responsible to forecast the future traffic condition.
    \remove{Finally, we describe how this framework is employed in order to answer travel time estimation queries at a city-scale.}
\end{abstract}



\maketitle

\section{Introduction}

In recent years, the wide usage of mobile devices and the corresponding collection of vast amounts of spatiotemporal data have resulted in the development of various novel Location Based Services (LBS). The LBS are software services that integrate geographic information providing  appropriate services and information to the users~\cite{schiller2004location}. Traffic forecasting and travel time estimation are undoubtedly two of the widely used LBS and a lot of recent research work has been conducted towards improving their performance. The importance of such services is indicated by the fact that the vast majority of drivers consults several times a day services that perform travel time estimation in order to appropriately choose the fastest route to follow.

\begin{figure}[t]
    \centering
    \includegraphics[width=0.47\textwidth]{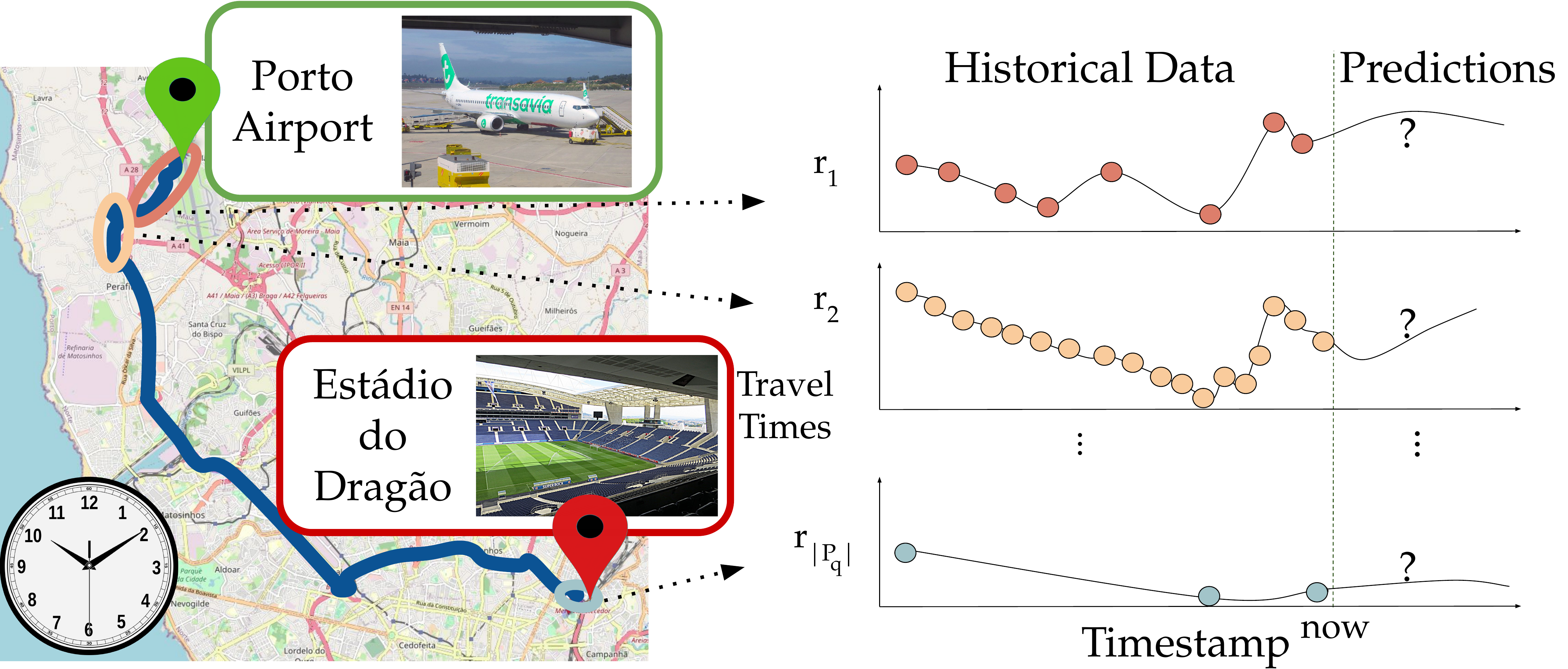}
    \caption{
    The travel time estimation problem for a given query path $P_q$ (blue line) and time of departure $t_q$ in the city of Porto, that starts at \textit{10:00} from the airport of Porto and ends at the Estádio do Dragão, the entire path is decomposed by a set of $|P_q|$ road segments $r_1\rightarrow r_2 \rightarrow \dots\rightarrow r_{|P|}$ and for each road segment we have a time series of travel time reports, received by the available probe vehicles.
    }
    \label{fig:intro}
\end{figure}

\remove{Several large corporations like Google Maps and TomTom perform accurate travel time estimation taking advantage of the voluminous amounts of spatiotemporal data that they receive in real time.
The use of such applications could not be universally applicable in every use case.
For instance, buses usually move in separate lanes from the private vehicles and make consecutive stops. 
In such cases services like Google Maps could not be queried in order to estimate the buses' arrival time, since such services collect information from private vehicles that move at different speeds without making consecutive stops.
Thus, it is critical to develop novel travel time estimation techniques that could be efficiently applied in cases where the GPS positions of a fleet with a limited number of probe vehicles are available.
}

Motivated by this, in this paper we propose a novel path based travel time estimation technique that considers the available traffic reports that have been received by the set of the available probe vehicles. Each probe vehicle moves in the road network and reports the time that was required to traverse each individual road segment. In this way, for each road segment of the road network a time series of the reported travel times are generated, illustrated at the right part of Figure~\ref{fig:intro}.
Our technique receives a query path along with a time of departure and estimates the time of arrival considering the current traffic condition of the road network.
Our problem is illustrated in Figure~\ref{fig:intro}. A query path $P_q$ and a time of departure $t_q$ are received as input and the task is to estimate the time that is required to traverse the whole path $P_q$ if the driver departs at $t_q$.


\remove{The design of a traffic forecasting technique for our setting, raises five main challenges, addressed in this paper:}

We propose a novel deep learning framework which is comprised of a recurrent part and a decoder. The recurrent part encodes the sparse traffic reports that are available at each time window using an attention mechanism and an embedding representations for each road segment. The decoder is responsible to forecast the traffic condition of the next time window. 

\remove{The rest of the paper is organized as follows. The related work
is discussed in Section~\ref{sec:related}. The formal problem definition along with our proposed technique are described in Section~\ref{sec:approach}. The evaluation of our technique is presented in Section~\ref{sec:experiments}. Finally, this work is concluded in Section~\ref{sec:conclusion}.
}

\section{Related Work}
\label{sec:related}

In \textit{DeepGTT} the travel time
distribution for any route is learnt by conditioning on the real-time traffic~\cite{li2019learning}. Initially, an embedding is estimated for each link considering its characteristics, then a nonlinear factorization model generates the speed and finally an attention mechanism is used to generate the observed travel time.
Also, in \textit{HETETA}~\cite{hong2020heteta} the road map is translated into a multi-relational network, considering the traffic behavior patterns. Temporal and graph convolutions are used in order to learn spatiotemporal heterogeneous information, considering recent, daily and weekly traffic.
\textit{CompactETA}~\cite{fu2020compacteta} provides an accurate ETA estimation with low latency. Graph attention network was employed in order to encode spatial and temporal dependencies of the weighted road network and the sequential information of the route is encoded with positional encoding. A multi-layer perceptron was used for online inference.
The authors in~\cite{li2018multi} proposed a multitask representation learning model which predicts the travel time of an origin-destination pair extracting a representation that preserves trips properties and road network structure.
\textit{ConSTGAT}~\cite{fang2020constgat} proposed a spatiotemporal graph neural network exploiting the spatial and temporal information with a 3D-attention mechanism and a model with convolutions over local windows in order to capture route's contextual information.
\textit{STGNN-TTE}~\cite{jin2022stgnn} adopted a spatial–temporal module to capture the real-time traffic condition and a transformer layer to estimate the links’ travel time and the total routes’ travel
time synchronously.

\section{Our Approach}
\label{sec:approach}


\subsection{Problem Definition}
\label{sec:prob_def}

\noindent\textbf{Road Network} is represented as a directed graph $G(V,E)$, where the nodes $V$ represent the junctions and the edges $E$ represent the $|E|$ roads segments. A road segment $r\in E$ is the part of the road network between two consecutive junctions without any intermediate junction between them.

\noindent\textbf{Trip} $T$ is a time ordered sequence of $|T|$  points $p_1 \rightarrow \dots \rightarrow p_{|T|}$; each point $p$ contains the geospatial coordinates of the moving object along with the corresponding timestamp $\tau$ that the vehicle was at this particular location $p=(lon, lat,\tau)$.

\noindent\textbf{Map-matched Trip} 
$T_G$ is a sequence of $|T_G|$ consecutive points $p'_1 \rightarrow \dots \rightarrow p'_{|T_G|}$ that comes from map matching trip $T$ on the road network $G$.  Each point $p'$ corresponds to a road segment that was traversed by $T$. 
Each point $p'$ of the map matched trip contains a triplet $(r,tt,\tau)$; $r$ is the traversed road segment, $tt$ is the \textbf{t}ravel \textbf{t}ime of the road segment $r$ and is computed assuming that the vehicle moved with the same speed in the road network between two consecutive GPS points and $\tau$ is the timestamp that the travel time is reported to the system.

\noindent\textbf{Travel time reports} $D$ is the collection of travel times for the road segments as they are extracted by the trips of all the available probe vehicles that traverse the road network. Each travel time report  $(r,tt,t, T_{id})$ contains the information of the map-matched trips enriched by the id of the trip $T_{id}$.

\noindent\textbf{Path} $P$ is a sequence of $|P|$ consecutive road segments $r_1 \rightarrow\dots\rightarrow r_{|P|}$, where $r_i$ is the $i$\textsuperscript{th} road segment of $P$.

Below we define formally the traffic forecasting problem.

\noindent\textbf{Traffic Forecasting:} \textit{ Given the available travel times of the last $L$ time windows $\mathcal{T}_{t-L+1:t}$ a traffic forecasting model forecasts the travel times of the next $H$ time windows $\mathcal{T}_{t+1:t+H}$, where the vector $\mathcal{T}_{t}$ contains the travel times of the $E$ road segments at time $t$. The input matrix $\mathcal{T}_{t-L+1:t}\in \mathbb{R}^{|E|\times L}$ has missing values for the roads that were not traversed by any vehicle at a given time window. The forecasted matrix $\mathcal{T}_{t+1:t+H}\in \mathbb{R}^{|E|\times H}$ contains forecasts for \textbf{all} the road segments $E$ for the next $H$ time windows.
}


\subsection{Data Preparation}
\label{sec:data_prep}
\begin{figure}
  \includegraphics[width=0.45\textwidth]{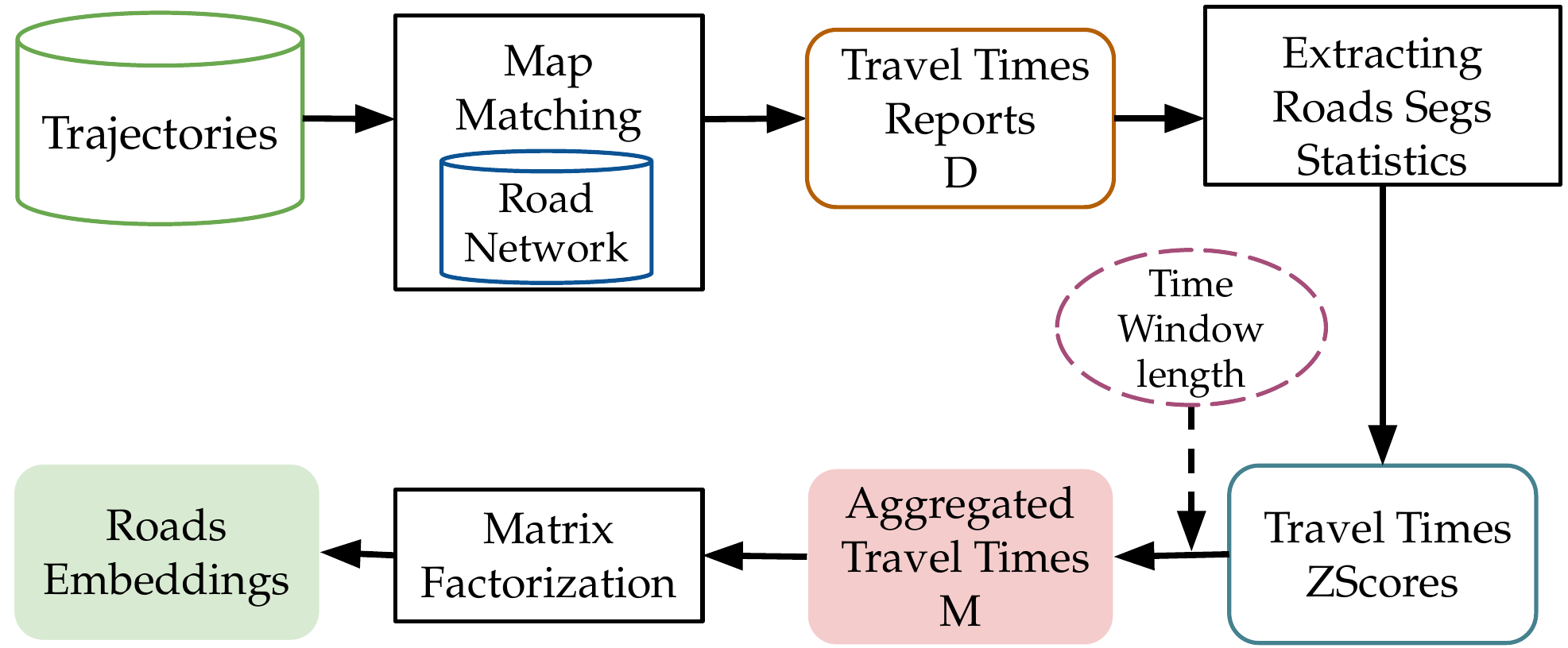}
  \caption{Data preparation.}
  \label{fig:datapreparation}
\end{figure}

The first step of the proposed framework is to preprocess the raw data and prepare them appropriately in order to feed them to the neural network. The overview of the data preparation approach is illustrated in Figure~\ref{fig:datapreparation} and described below.

\noindent\textbf{Map Matching.}
Firstly, we map-match the available trips matching them to the road network $G$. Each trip $T$ is transformed into a map-matched trip $T_G$. This procedure generates the set of the available travel time reports $D$.  This step is common to both the historical data that are used to train our model and the streaming traffic data that will be used to make forecasts in real time.

\noindent\textbf{Modeling the periodicity of traffic.} In order to model the periodicity of traffic we estimate from the historical travel time reports the average travel time $avg\_tt_{i,hour}$ for each road segment $r_i\in E$ and for different hours of day $hour\in[1\dots24]$. Then, we subtract from each travel time the historical average travel time for that road segment at the given hour. In this way, we force the deep learning framework to model, for each different road segment, the deviation from the average travel time for the different hours of the day.

\noindent\textbf{Standardizing Travel Times.}
Since road segments have different lengths and speed limits we selected to standardize the travel time reports, considering the average behaviour of  each different road segment. More specifically, for each road segment $r_i$ we compute the historical average travel time $\mu_i$ and standard deviation of travel times $\sigma_i$ and we use these values in order to standardize the travel times per road segment. For instance, if $tt_5$ is a travel time that is reported for the road segment $r_5$ then the corresponding Z-Score will be $\frac{tt_5-\mu_5}{\sigma_5}$.  In the rest of the paper we assume that travel times are the Z-Scores of travel times with subtracted the average historical travel time for the different hours of the day.

\noindent\textbf{Aggregating travel times.}
The historical travel time reports $D$ are grouped together generating a sparse matrix $M\in\mathbb{R}^{|E|\times W}$.  The rows of $M$ correspond to the $|E|$ road  segments of the road network $G$ and the columns correspond to the $W$ time windows. In this work we use time windows of $15$ minutes. 
If more than one travel time reports are available for a particular road segment $r_i$ at the same time window $w_j$ then $M_{ij}$ contains the average travel time of the available travel times.

\begin{figure*}[t]
\centering
\begin{minipage}{.33\textwidth}
  \centering
  \includegraphics[width=0.99\textwidth]{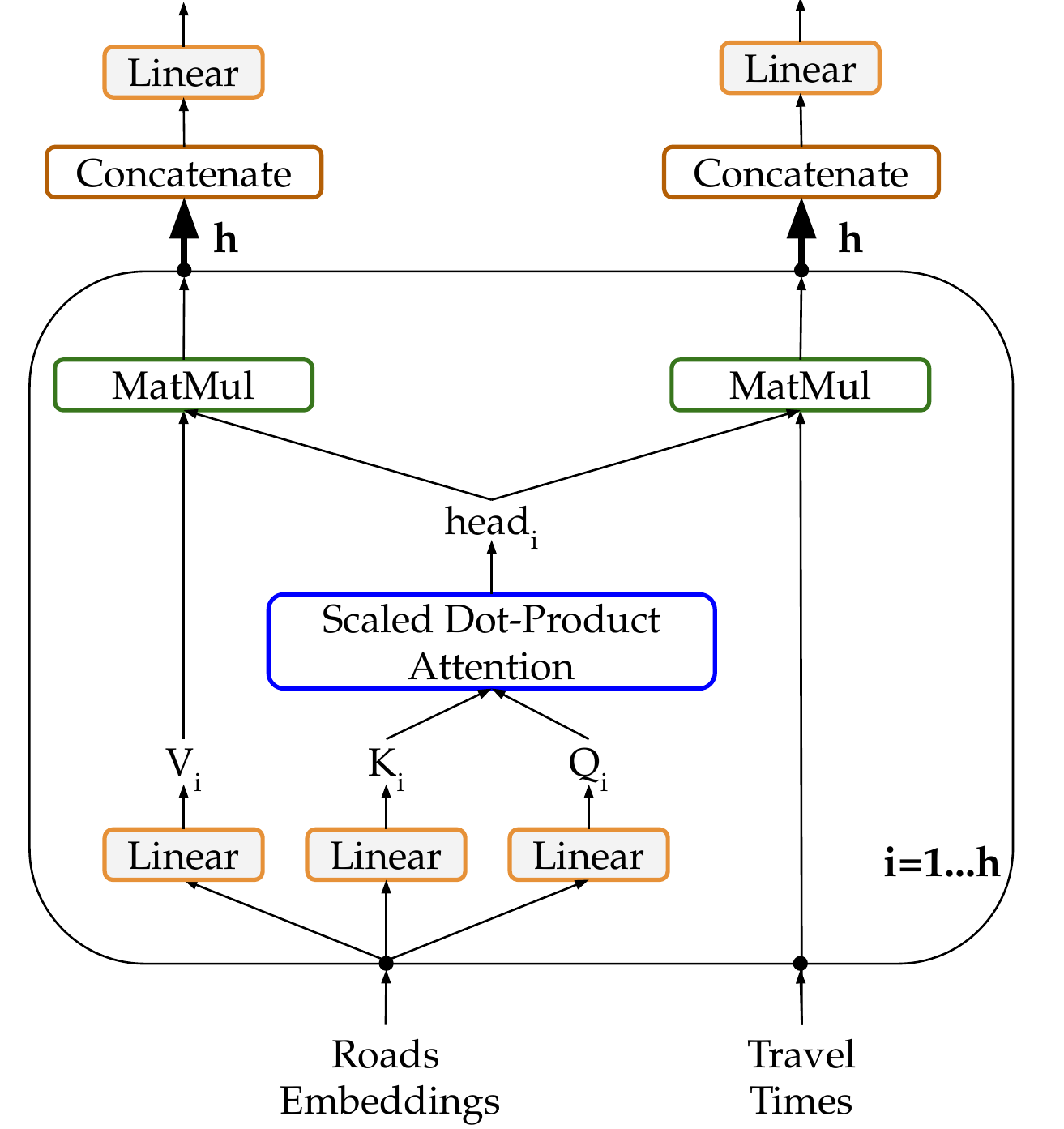} \caption{Multi-Head Scaled Dot-Product Attention}
  \label{fig:attention_overview}
\end{minipage}%
\begin{minipage}{.3\textwidth}
  \centering
  \includegraphics[width=0.99\textwidth]{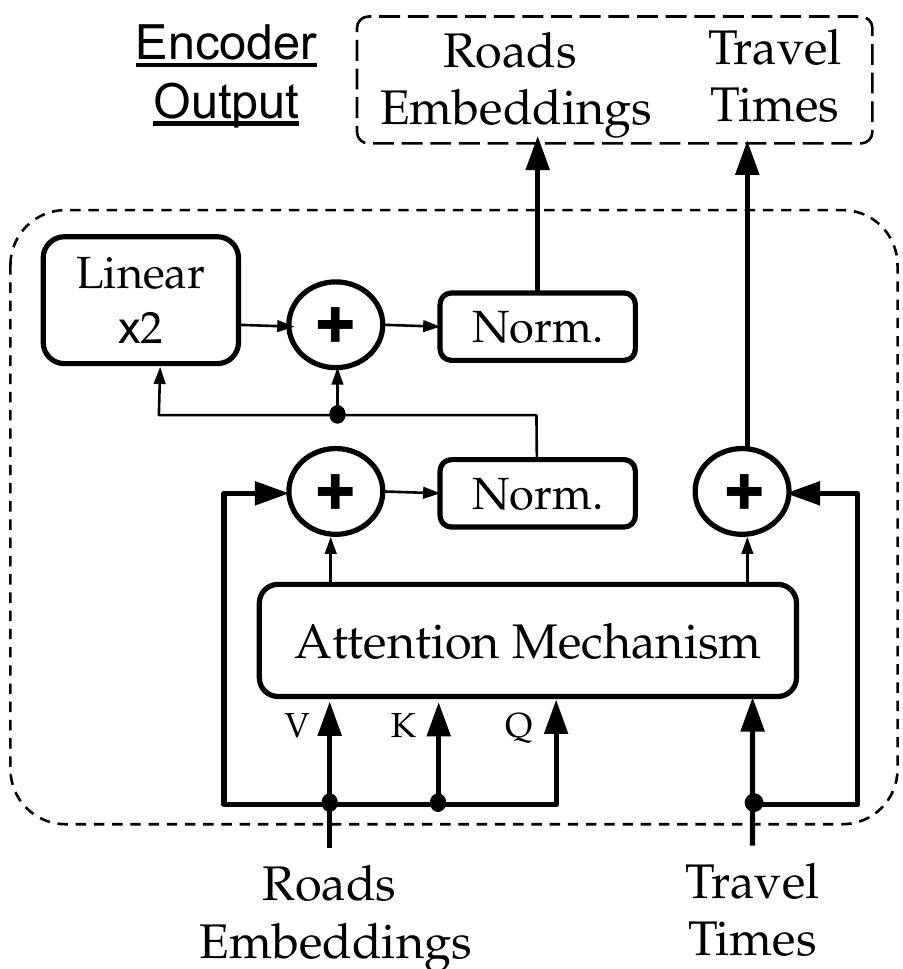}
  \caption{Encoder Block.}
  \label{fig:encoder}
\end{minipage}%
\begin{minipage}{.33\textwidth}
  \centering
  \includegraphics[width=0.99\textwidth]{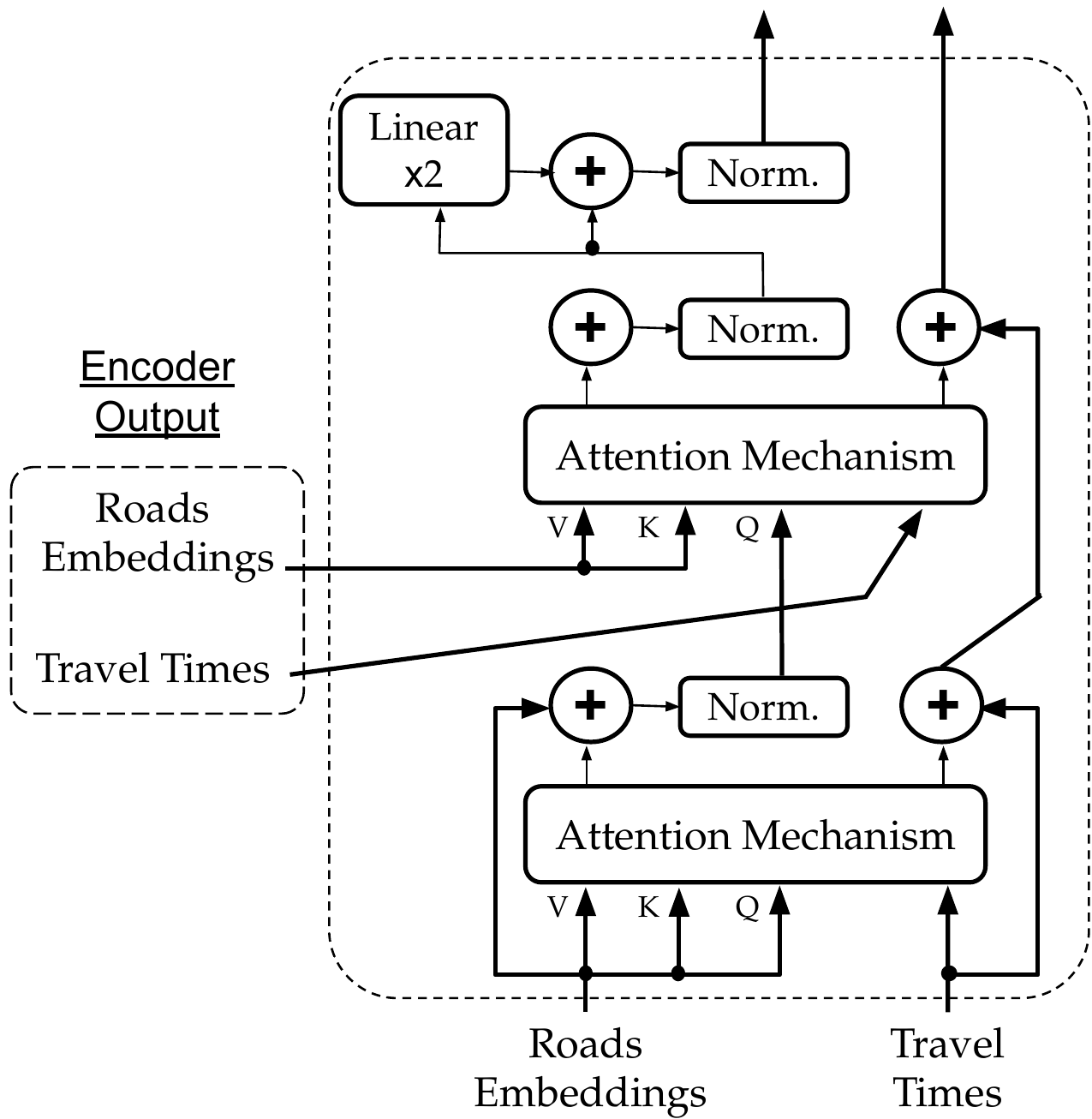}
  \caption{Decoder Block.}
  \label{fig:decoder}
\end{minipage}%
\end{figure*}

\noindent\textbf{Extracting Road Segments Embeddings.} An embedding representation $E_i$ is detected for each road segment $r_i$ considering its historical travel time reports. Here, we follow the process introduced by~\cite{zygouras2019htte}. We perform matrix factorization in the sparse matrix $M$,
learning a matrix $\mathcal{P} \in \mathbb{R}^{|E|\times d}$ contains a $d$-dimensional embedding representation of the available road segments


\noindent\textbf{Feeding the Model.}
The deep learning model that is described in Section~\ref{sec:architecture} receives as input two vectors that contain: \textit{(i)} the aggregated travel times that are available for a given time window and \textit{(ii)} the corresponding road segments. 
For instance, consider a road network $G$ that is comprised of $|E| = 5$ road segments $[r_i]$, $i\in[1,\dots,5]$. If for a particular time window only the travel times $tt_2$ and  $tt_5$ of road segments $r_2$ and $r_5$ respectively are available, then the inputs to the deep learning model will be the following: the vector of the travel times $\mathcal{T} = [tt_2,tt_5,\emptyset,\emptyset,\emptyset]\in \mathbb{R}^{|E|}$ and  the vector of the road segments ids $\mathcal{R} = [r_2,r_5,\emptyset,\emptyset,\emptyset]\in \mathbb{Z}^{|E|}$. 
Then, inside the deep learning model the ids of the road segments are transmitted to an embedding layer. This layer transforms the vector $\mathcal{R}$ into a matrix of road segments embeddings $\mathcal{E} = [E_2,E_5,\emptyset,\emptyset,\emptyset]\in \mathbb{R}^{|E|\times d}$. The embedding representation of the road segments is  trainable and initialized with matrix $\mathcal{P}$, computed using matrix factorization as it was described above.

\subsection{Attention Mechanism}
\label{sec:attention_mechanism}


Here, we extend the "Scaled Dot-Product Attention" that was introduced in \cite{vaswani2017attention}. The proposed attention mechanism encodes a varying number of travel time reports received at a particular time window. We consider as input here the following: 
\textit{(i)} the embeddings' matrix of the road segments that have been traversed by the probe vehicles at a particular time window along with
\textit{(ii)} the vector of the corresponding reported travel times.
The overview of the proposed attention mechanism is illustrated in Figure~\ref{fig:attention_overview}.

Initially, the Query $Q_i$, Key $K_i$ and Value $V_i$ matrices are computed using the embeddings $\mathcal{E}$ of the available road segments, computed earlier. Therefore, three parameter matrices
$W_i^Q\in\mathbb{R}^{d\times d}$,
$W_i^K\in\mathbb{R}^{d\times d}$ and
$W_i^V\in\mathbb{R}^{d\times d}$
are trained using the training instances and are used to compute the matrices $Q_i=\mathcal{E}W_i^Q$, $K_i=\mathcal{E}W_i^K$ and $V_i=\mathcal{E}W_i^V$. The index $i\in[1,\dots,h]$ of the different parameter matrices stands for the $h$ parallel attention layers.

The next step is to compute the attention scores using the $Q_i$ and $K_i$ matrices. The scores indicate the focus that will be placed at the travel times of other road segments, that have been reported at the same time window. Multiple attention heads $head_i \in\mathbb{R}^{|E|\times|E|}$ are computed in parallel according to eq.~\ref{eq:scaled_dot_product_att} indicating the attention at each particular road segment.

\begin{equation}
    head_i=softmax(\frac{Q_iK_i^T}{\sqrt{d}}),\, i\in[1,\dots,h]
    \label{eq:scaled_dot_product_att}
\end{equation}

The road segments' embeddings and  travel times are then updated considering the computed attention heads. More specifically we train the parameter matrices
$W^{O_1}\in\mathbb{R}^{hd\times d}$ and $W^{O_2}\in\mathbb{R}^{h|E|\times |E|}$ that are multiplied with the concatenated values $V_i$ and travel times $\mathcal{T}$ respectively.

\begin{equation}
    \mathcal{E}' = Concat(head_1V_1,\dots,head_hV_h)W^{O_1}
\end{equation}

\begin{equation}
    \mathcal{T}' = Concat(head_1\mathcal{T},\dots,head_h\mathcal{T})W^{O_2}
\end{equation}

\subsection{Traffic Transformer's Architecture}
\label{sec:architecture}

Here we describe our model's architecture, which is based on the original implementation of Transformer model described in~\cite{vaswani2017attention}. 

\subsubsection{Encoder}
The encoder considers all travel time reports that are available at a given time window, encodining the traffic condition of that time window. It is comprised by a set of $N$ identical blocks. The first block receives as input the roads segments embeddings and the travel times that are available at a given time window, following the data preparation procedure described in Section~\ref{sec:data_prep}. The rest encoder blocks receive as input the output of the previous block. Figure~\ref{fig:encoder} illustrates the overview of the encoder block.

Each block first transmits the matrix of the available roads embeddings $\mathcal{E}$ and the corresponding vector of travel times $\mathcal{T}$ at the attention mechanism. The attention mechanism produces the matrix $\mathcal{E}'$ and the vector $\mathcal{T}'$. Then, residual connections are employed at the output of the attention mechanism, normalizing the sum of the received roads segments embeddings $\mathcal{E}$ with the output of the attention mechanism $\mathcal{E}'$. The output is transmitted to two dense layers followed by another residual connection. For the travel times the output of each encoder block is the sum of the received travel times $\mathcal{T}$ and the output of the attention mechanism $\mathcal{T}'$.


\subsubsection{Decoder} 




\begin{figure}
    \centering
    \includegraphics[width=.99\linewidth]{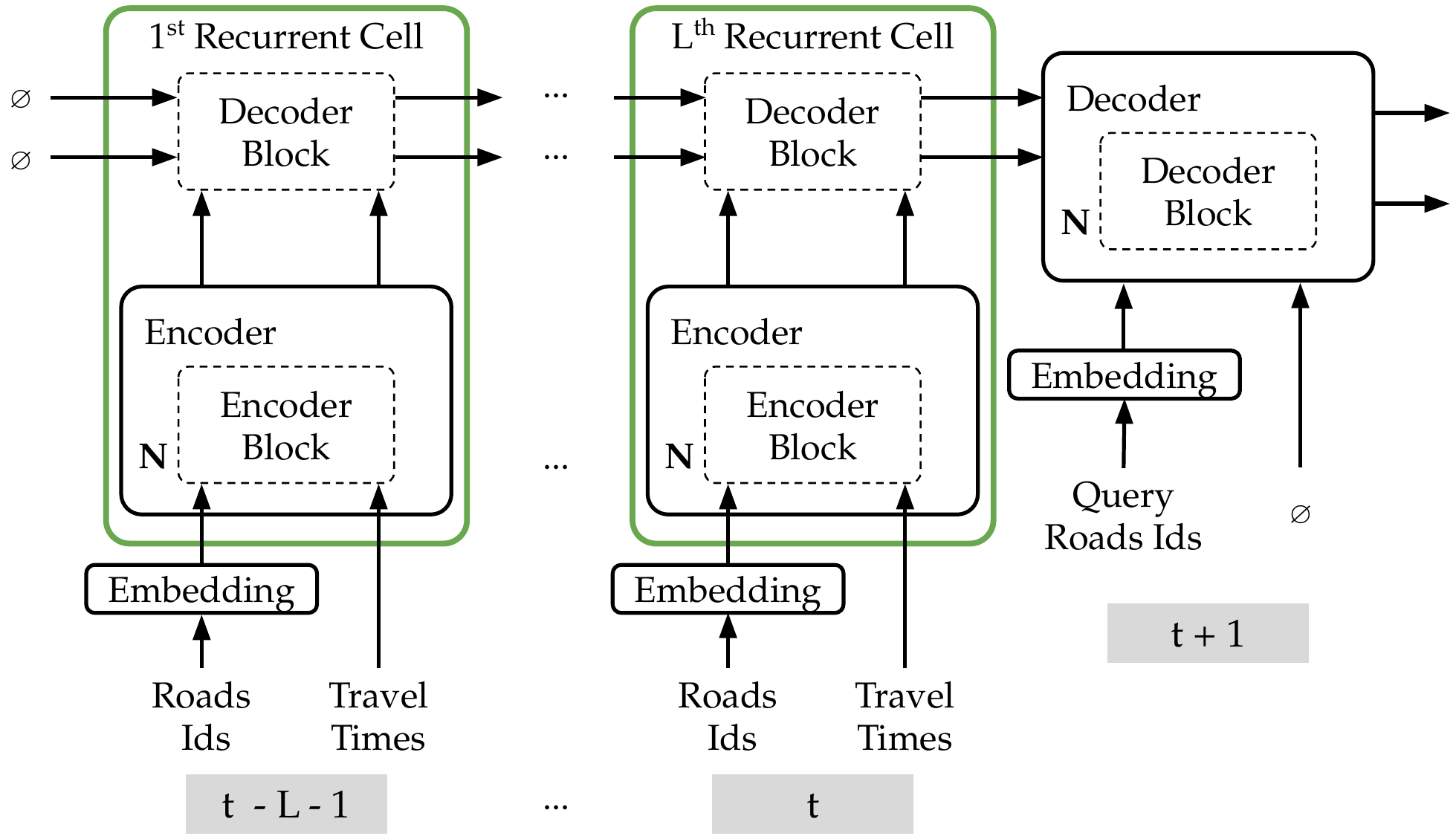}
    \captionof{figure}{Overview of our model.}
    \label{fig:recurrent_overview}
\end{figure}

The decoder (Figure~\ref{fig:decoder}) is responsible to forecast the travel times of the next time window, considering the encoder's output. The decoder consists of a set of $N$ blocks, similarly to the encoder.
Each block receives as input the output of the encoder and the output of the previous block. In the training phase the first block receives as input \textit{(i)} the embeddings of the road segments that are available in the target time window and \textit{(ii)} a vector of zeros. For the testing phase the first block receives as input \textit{(i)} the embeddings of all the road segments $E$ and \textit{(ii)} a vector of zeros. Recall that we are working with the Z-Scores of travel times aggregated per road segment. Consequently, the vector of zeros corresponds to the average travel for each road segment. 

Each block of the decoder contains two attention mechanisms. Firstly, the embeddings of the queried road segments and the travel times are transmitted to the first attention mechanism, followed by a residual connection. Then a second attention mechanism is employed, receiving as input the embedding matrix $\mathcal{E}_1'$ that resulted from the first attention mechanism along with the embeddings and the travel times that come from the output of the encoder. The main difference here is that the matrices $V_i$ and $K_i$ are computed from the output of the encoder and that the considered travel times come from the encoder. Then, the embedding output $\mathcal{E}_2'$ of the second attention mechanism is followed again by a residual connection. This is followed by two dense layers and a second residual connection. Finally, the travel times that result from each block is the sum of the original travel times $\mathcal{T}$ that were received as input along with travel times that result from the first and the second attention mechanism $\mathcal{T}_1'$  and $\mathcal{T}_2'$ respectively.

\subsubsection{Recurrent Neural Network}

The final module of our proposed model is a recurrent model that considers the sequence of the last $L$ time windows. Each cell of the recurrent network encapsulates an encoder (consisting of $N$ encoder blocks) along with a single decoder block. Here the decoder block is responsible to aggregate the information that has been encoded from the previous time window with the information that has been encoded from the current time window. Figure~\ref{fig:recurrent_overview} illustrates this recurrent architecture. The encoder and the decoder blocks of the different recurrent cells share the same weights among the $L$ different time windows.

The output of the last recurrent cell is used by the decoder model in order to make forecasts. The decoder model consists of $N$ decoder blocks that are different from each other and from the decoder block that lies inside the recurrent cells. The output of the last decoder block contains the predicted travel times of the queried road segments for the next time window. This will be the Z-Scores of the travel times for the road segments that were queried  at the first decoder block.

\section{Conclusion}
\label{sec:conclusion}
In this paper we presented a novel deep learning framework that considers the current traffic condition of the road network and is used to forecast the traffic condition.  Our framework can efficiently encode the travel time reports that are available at a particular time window via an attention mechanism that considers only the available travel times reports and the corresponding embeddings of the road segments. 

\begin{acks}
This research has been financed by the European Union through the H2020 LAMBDA Project (No. 734242),
the EU ICT-48 2020 project TAILOR (No. 952215) and the Horizon Europe AUTOFAIR Project (No. 101070568).

\end{acks}


\printbibliography

@String{Computing = "Computing" }

@String{Computer = "{IEEE} Computer" }

@ArtifactSoftware{R,
    title = {R: A Language and Environment for Statistical Computing},
    author = {{R Core Team}},
    organization = {R Foundation for Statistical Computing},
    address = {Vienna, Austria},
    year = {2019},
    url = {https://www.R-project.org/},
}

@article{vaswani2017attention,
  title={Attention is all you need},
  author={Vaswani, Ashish and Shazeer, Noam and Parmar, Niki and Uszkoreit, Jakob and Jones, Llion and Gomez, Aidan N and Kaiser, {\L}ukasz and Polosukhin, Illia},
  journal={Advances in neural information processing systems},
  volume={30},
  pages={5998--6008},
  year={2017}
}

@inproceedings{li2018multi,
  title={Multi-task representation learning for travel time estimation},
  author={Li, Yaguang and Fu, Kun and Wang, Zheng and Shahabi, Cyrus and Ye, Jieping and Liu, Yan},
  booktitle={Proceedings of the 24th ACM SIGKDD International Conference on Knowledge Discovery \& Data Mining},
  pages={1695--1704},
  year={2018}
}

@book{schiller2004location,
  title={Location-based services},
  author={Schiller, Jochen and Voisard, Agn{\`e}s},
  year={2004},
  publisher={Elsevier}
}

@inproceedings{zygouras2019htte,
  title={HTTE: A Hybrid Technique For Travel Time Estimation In Sparse Data Environments},
  author={Zygouras, Nikolaos and Panagiotou, Nikolaos and Li, Yang and Gunopulos, Dimitrios and Guibas, Leonidas},
  booktitle={Proceedings of the 27th ACM SIGSPATIAL International Conference on Advances in Geographic Information Systems},
  pages={99--108},
  year={2019}
}

@inproceedings{li2019learning,
  title={Learning travel time distributions with deep generative model},
  author={Li, Xiucheng and Cong, Gao and Sun, Aixin and Cheng, Yun},
  booktitle={The World Wide Web Conference},
  pages={1017--1027},
  year={2019}
}

@inproceedings{hong2020heteta,
  title={Heteta: heterogeneous information network embedding for estimating time of arrival},
  author={Hong, Huiting and Lin, Yucheng and Yang, Xiaoqing and Li, Zang and Fu, Kung and Wang, Zheng and Qie, Xiaohu and Ye, Jieping},
  booktitle={Proceedings of the 26th ACM SIGKDD International Conference on Knowledge Discovery \& Data Mining},
  pages={2444--2454},
  year={2020}
}

@inproceedings{fu2020compacteta,
  title={Compacteta: A fast inference system for travel time prediction},
  author={Fu, Kun and Meng, Fanlin and Ye, Jieping and Wang, Zheng},
  booktitle={Proceedings of the 26th ACM SIGKDD International Conference on Knowledge Discovery \& Data Mining},
  pages={3337--3345},
  year={2020}
}

@inproceedings{fang2020constgat,
  title={Constgat: Contextual spatial-temporal graph attention network for travel time estimation at baidu maps},
  author={Fang, Xiaomin and Huang, Jizhou and Wang, Fan and Zeng, Lingke and Liang, Haijin and Wang, Haifeng},
  booktitle={Proceedings of the 26th ACM SIGKDD International Conference on Knowledge Discovery \& Data Mining},
  pages={2697--2705},
  year={2020}
}

@article{jin2022stgnn,
  title={STGNN-TTE: Travel time estimation via spatial--temporal graph neural network},
  author={Jin, Guangyin and Wang, Min and Zhang, Jinlei and Sha, Hengyu and Huang, Jincai},
  journal={Future Generation Computer Systems},
  volume={126},
  pages={70--81},
  year={2022},
  publisher={Elsevier}
}

\end{document}